\documentclass[10pt,twocolumn,letterpaper]{article}
\usepackage[pagenumbers]{cvpr} 

\usepackage[dvipsnames]{xcolor}

\usepackage{algorithmic}
\usepackage{algorithm}

\usepackage[accsupp]{axessibility} 
\definecolor{cvprblue}{rgb}{0.21,0.49,0.74}
\usepackage{hyperref}
\hypersetup{pagebackref,breaklinks,colorlinks,citecolor=cvprblue}

\def\paperID{10}
\def\confName{CVPR}
\def\confYear{2024}

\title{ImplicitTerrain: a Continuous Surface Model for Terrain Data Analysis}

\author{Haoan Feng,  Xin Xu,  Leila De Floriani\\
University of Maryland, College Park\\
{\tt\small \{hfengac, xinxu629, deflo\}@umd.edu}
}

\begin{document}
\maketitle
\begin{abstract}
    Digital terrain models (DTMs) are pivotal in remote sensing, cartography, and landscape management, requiring accurate surface representation and topological information restoration. While topology analysis traditionally relies on smooth manifolds, the absence of an easy-to-use continuous surface model for a large terrain results in a preference for discrete meshes. Structural representation based on topology provides a succinct surface description, laying the foundation for many terrain analysis applications. However, on discrete meshes, numerical issues emerge, and complex algorithms are designed to handle them. This paper brings the context of terrain data analysis back to the continuous world and introduces ImplicitTerrain\footnote{Project homepage available at~\url{https://fengyee.github.io/implicit-terrain/}}, an implicit neural representation (INR) approach for modeling high-resolution terrain continuously and differentiably. Our comprehensive experiments demonstrate superior surface fitting accuracy, effective topological feature retrieval, and various topographical feature extraction that are implemented over this compact representation in parallel. To our knowledge, ImplicitTerrain pioneers a feasible continuous terrain surface modeling pipeline that provides a new research avenue for our community.

\end{abstract}
\section{Introduction}
\label{sec:intro}

In geographic information systems and remote sensing, Digital Terrain Models (DTMs) have emerged as foundational elements for a myriad of applications spanning environmental management, urban planning, disaster control, and beyond. Morse theory~\cite{milnor1963morse} is one of the fundamental mathematical tools for exploring topological features that supports various downstream applications shape segmentation~\cite{de2013discrete,xu2023topology} to road network analysis~\cite{dey2017improved}. For a terrain height field, Morse theory interprets the terrain surface as a smooth manifold. However, obtaining a continuous surface model through interpolating for large-scale terrain data confronts formidable challenges: prohibitive computational cost, sophisticated surface modeling depending on hyperparameter tuning, and a lack of topological integrity. Consequently, continuous surface models find limited utility in topological analysis but are used for topography and visualization purposes~\cite{mitavsova1993interpolation}. In contrast, discrete mesh representations of the terrain surface are more commonly used not only for topological analysis but for terrain processing, visualization, and analysis~\cite{scoville2019discrete,de2015morse}. However, the discrete mesh resolution is limited by the computational resources. Discrete methods usually suffer from rough surface approximation and numerical instability.~\cite{de2015morse}.
\begin{figure}[t]
    \centering
    \includegraphics[width=0.9\columnwidth]{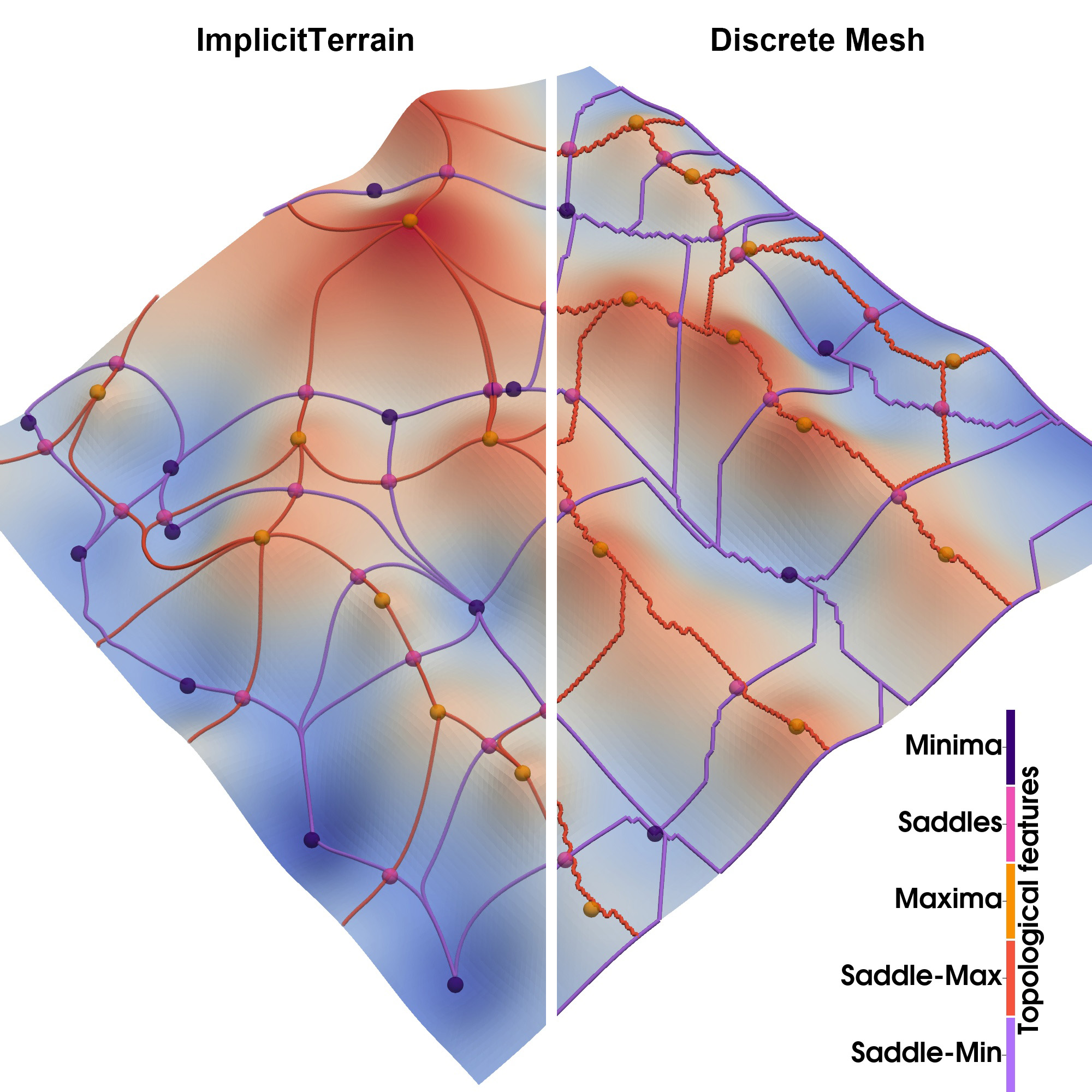}
    \caption{A 3D view of topological features derived from ImplicitTerrain's smooth surface model (left) and discrete mesh model (right) over a synthetic terrain dataset. Different critical point types are highlighted in different colors. Separatrix lines from our ImplicitTerrain are smoothly aligned with the terrain surface and color-coded by the critical point pair they connected.}
    \label{fig:f1}
\end{figure}
There is a need for a continuous surface model that can be used for topological analysis and is scalable to large-scale terrain data. Such accuracy of terrain surface and its high-order derivatives is pivotal for topological feature extraction and enhances many other analyses of terrain data, such as topography, hydrology, and geomorphology. In this paper, we propose a novel continuous surface modeling pipeline, ImplicitTerrain, that leverages the recent advances in the implicit neural representation (INR) to achieve a continuous surface model for terrain data analysis. For example, in~\cref{fig:f1}, topological features extracted from ImplicitTerrain are well-aligned with the synthetic terrain surface. Benefiting from the interpolation capability, INR is a revolutionary representation that may achieve better memory efficiency than a discrete representation by decoupling the network size with the input data size.

However, there is still a gap between practical applications and the theoretical potential of INR. INR's ``over-smoothness'' fitting result makes reconstructing the detailed surface geometry difficult, especially for high-resolution input~\cite{Rahaman2018OnTS}. Despite advancements (\eg,~\cite{sitzmann2020implicit,tancik2020fourier}), for visual signals consisting of complex frequency components, obtaining accurate high-order derivatives is still challenging. The lengthy training process and the high computational cost of INR also limit its application in large-scale terrain data fitting and analysis. Besides, there is a domain difference between fitting the terrain height field and the photometric data, that the former requires both a plausible visual appearance and accurate surface restoration. Merging the gap between the theoretical potential and the practical applications of INR, ImplicitTerrain pioneers a feasible continuous terrain surface modeling pipeline that opens a new research avenue. Our main contributions are as follows:
\begin{itemize}
    \item We introduce a novel Surface-plus-Geometry (SPG) cascaded INR model for high-resolution terrain surface modeling that keeps high reconstruction fidelity, reduces overall training time, and allows the topological analysis to be directly performed on the continuous manifold.
    \item We propose a progressive training strategy to train the SPG model from coarse to fine, significantly improving convergence speed while fulfilling the pre-processing requirement of topological and topographical analysis.
    \item Beyond surface fitting, we demonstrate the good alignment of the extracted topological features with the results based on the discrete Morse theory. Moreover, it shows superior perceptional and visual quality, reflecting terrain surfaces as perceived by human beings.
    \item Demonstrating ImplicitTerrain's utility in calculations of various topographical features, \ie, slope, aspect, and curvature, directly from surface derivatives, underscoring its potential for comprehensive terrain data analysis.
\end{itemize}

\section{Background and Related Work}\label{sec:relatedwork}
\subsection{Implicit neural representation}
Different from the widely-known deep neural networks aiming to learn the hidden high-dimensional features such as semantic information of visual signal, INR is designed for reconstructing the input through neural network weights and it can be regarded as a revolutionary representation of the visual signals~\cite{xie2022neural}. Most INRs are rooted in the Universal approximation theorem~\cite{hornik1991approximation} - for a multi-layer perceptron (MLP) with a sufficient number of hidden units, any field function can be approximated within any precision~\cite{chen2019learning}. For an MLP $\Psi\colon \mathbb{R}^d \to \mathbb{R}^n$ to reconstruct a visual signal, a common design of the network is to model a regression problem, in which the network takes individual coordinates $\vec{x} \in \mathbb{R}^d$ as input and predicts the field function values $\mathcal{F} |_{\vec{x}} \in \mathbb{R}^n$ corresponding to the coordinates, \ie, $\Psi(\vec{x}) \approx \mathcal{F} |_{\vec{x}}$. Take a RGB image as an example, a neural work takes pixel coordinates $(u,v) \in \mathbb{R}^2$ and outputs pixel color $(r,g,b) \in \mathbb{R}^3$. This design of the network is also named as \textit{coordinate-based neural network}. INR converts the signal to a continuous function representation defined on the input domain. Consequently, INR can be inferred in arbitrary resolutions and achieve better memory efficiency than a discrete representation by decoupling the network model size with the input data size~\cite{sitzmann2020implicit}. The network can be stochastically optimized to be efficient and scalable to large input data~\cite{atzmon2020sal}. Apart from the inherent continuity, INR is differentiable by an automatic differentiation technique that allows function gradient and higher-order derivatives to be easily accessible at any query location~\cite{sitzmann2020implicit}.

INR has gained much attention in various research fields where 2D images~\cite{chen2021learning,paz2022multiresolution,luo2023sr}, 3D shapes~\cite{niemeyer2020differentiable,genova2020local,lin2020sdf,xu2019disn,chibane2020implicit,wang2021neus,park2019deepsdf,atzmon2020sal,gropp2020implicit}, textures\cite{saito2019pifu,chen2019learning,oechsle2019texture,henzler2020learning}, light fields~\cite{derksen2021shadow,bemana2020x}, dynamic fields~\cite{du2021neural,jiang2022lord}, and indoor and outdoor scenes~\cite{sitzmann2019scene,niemeyer2021giraffe,stucker2022implicity,xiangli2022bungeenerf,jiang2020local,lindell2022bacon} are proposed to be converted into neural representations. For 3D objects, signed/unsigned distance functions~\cite{park2019deepsdf,chibane2020neural,chabra2020deep,yariv2021volume} and occupancy prediction networks~\cite{mescheder2019occupancy,peng2020convolutional} are mostly used, while for 2D images, single MLP~\cite{sitzmann2020implicit,fathony2020multiplicative} and decoder for feature grids~\cite{chen2021learning,muller2022instant} are more common. Recently, researchers have shown the representation power of INR in biomedical data~\cite{yang2023sci,molaei2023implicit}, satellite images~\cite{derksen2021shadow,stucker2022implicity,xiangli2022bungeenerf}, and physics-informed simulations~\cite{cuomo2022scientific,Chiu2021CANPINNAF}. Others take advantage of INR's compression capabilities in image~\cite{Strmpler2021ImplicitNR,yang2023sci,Catania2023NIFAF,Dupont2021COINCW} and video compression~\cite{Chen2021NeRVNR,Chen2022CNeRVCN}.

Through Fourier analysis, many neural networks, such as ReLU MLP, suffer from capturing high-frequency components of the input signals, because their fitting power biases toward the low-frequency components, known as {\em spectral bias}~\cite{Rahaman2018OnTS,Basri2020FrequencyBI,yuce2022structured}. It leads to the ``over-smoothness" issue. FFN~\cite{tancik2020fourier} makes use of Neural Tangent Kernel (NTK) techniques to analyze this phenomenon and proposes to use positional encoding techniques to counter the spectral biases and capture more high-frequency details. Another well-known approach is SIREN~\cite{sitzmann2020implicit}, which makes use of the sinusoidal non-linear activation function with carefully designed network initialization. A SIREN model can be expressed as
\begin{align}
    \Psi(\vec{x})   & = \Omega_n (\psi_{n-1} \circ \psi_{n-2} \circ \cdots \circ \psi_{0})(\vec{x}) + \vec{b}_n \\
    \psi_i(\vec{x}) & = \sin(\Omega_i \vec{x} + \vec{b}_i)
\end{align}
where $\psi_i$ is the $i$-th layer of the MLP and $(\Omega_i, \vec{b}_i)$ is the corresponding network weights and biases. This design choice allows the model to be written in a composition of sinusoidal functions and the derivatives of the network have similar expressions such that the training process is not only targeting the field function $\mathcal{F}$ but also its derivatives $\{\nabla_{\vec x}\mathcal{F}, \nabla_{\vec x}^2\mathcal{F},\ldots \}$~\cite{sitzmann2020implicit}. This design is important for topological analysis and topographical feature calculation since the reconstructed surface function is smooth and an accurate gradient field is necessary. ImplicitTerrain takes advantage of the compactness and the representation power of INR which may achieve better accuracy with less computational resources. As shown in this paper, ImplicitTerrain demonstrates the effectiveness and implementation simplicity in the field of terrain data analysis.

\subsection{Topology-based terrain analysis}
\label{sec:tda}
\textbf{Morse theory.} Morse theory captures the relationships between the topology of a manifold and the critical points of a function defined on it~\cite{milnor1963morse}. For a smooth scalar function $f$ on domain $D \subseteq \mathbb{R} ^2$, {\em critical points} of $f$ are those gradients of $f$ vanishes. The index $k_p$ of a critical point $p$ is the number of negative eigenvalues of the Hessian matrix $\mathcal{H}|_{p}$ of the second-order partial derivatives of $f$ at $p$. If $k_p=0$, $p$ is a minimum, $k_p=1$ a saddle, and $k_p=2$ a maximum. A smooth function $f$ is a {\em Morse function} if every critical point $p$ is non-degenerate, \ie, $\mathcal{H}|_{p}(f) \neq 0$.

An {\em integral line} of $f$ is a path tangent to gradient $\nabla f$, tracing the steepest ascent or descent from one critical point to another. An integral line connecting two critical points of consecutive index is called a {\em separatrix} line. Given a critical point $p$ of index $k$, the integral lines converging at $p$ of index $k_p$ form a $k$-cell, called the {\em descending $k$-manifold} of $p$. Dually, integral lines originating at $p$ form a ($d$-$k$)-cell, called the {\em ascending $k$-manifold} of $p$. The ascending manifolds partition $D$ into cells forming the {\em ascending Morse complex} and {\em descending Morse complex}, respectively. 

A Morse function $f$ is a {\em Morse-Smale} function when the ascending and descending manifolds intersect transversally. Cells obtained as the intersection of descending and ascending 2-manifolds of a Morse-Smale function $f$ decompose $D$ into a {\em Morse-Smale} complex. A {\em critical net} is the 1-skeleton of the Morse-Smale complex, which consists of the critical points and the separatrix lines connecting them. 

The combinatorial structure of the critical net is described by the {\em Morse Incidence Graph} (MIG)~\cite{comic2010building}. The nodes of the MIG correspond to the critical points of $f$, while its arcs correspond to the separatrix lines, \ie, the connections between pairs of critical points. A value is associated with each arc of the graph, which is the absolute value of the difference between the function value at the critical points corresponding to its extreme nodes, called {\em persistence}. Insignificant critical points that do not contribute to the main topological structure of a surface are usually removed by the MIG based on the persistence values. The two fundamental operators for simplifying a MIG, while maintaining its topological integrity are: {\em Minimum-Saddle-Minimum}, which collapses two minima adjacent to the same saddle into one of them, eliminating also the common saddle, and its dual {\em Maximum-Saddle-Maximum}, which is defined symmetrically~\cite{de2015morse}.

\textbf{Discrete approaches.} A variety of methods have been proposed in the literature to identify critical points and trace separatrix lines in the discrete case when a terrain is defined by a set of points either at the vertices of a regular square grid (gridded data), or scattered in the plane. In the case of gridded data, earlier approaches interpolate the data through smooth functions, including a $C1$-differentiable Bernstein-Bezier bi-cubic function~\cite{bajaj1998visualization}, as well as bilinear $C0$-interpolant and bi-quadratic interpolant~\cite{schneider2004construction, schneider2005extraction}, with derivatives computed numerically in~\cite{bajaj1998visualization} and analytically in~\cite{schneider2004construction, schneider2005extraction}. The efficacy of these methods depends on precise tuning of the approximation functions to avoid false critical points and maintain domain-wide continuity.

Smooth interpolation methods also struggle with scalability, especially with large terrain datasets. Thus, discrete methods based on triangulating the data points have been most widely used, for instance, in Topology ToolKit (TTK)~\cite{tierny2017topology}, a popular topological data analysis toolkit.

Two discrete approaches have been applied to triangulated terrains, namely {\em piecewise-linear Morse theory} by Banchoff~\cite{banchoff1967critical} and {\em discrete Morse theory} by Forman~\cite{forman1998morse}. The former identifies and classifies critical points per~\cite{banchoff1967critical} for polyhedral surfaces, tracing separatrix lines from saddles to minima and maxima along triangle edges.

Discrete Morse theory on a triangle mesh is based on extending the elevation function from the vertices to the edges and triangles of the mesh and on defining a discrete gradient field, usually called a {\em Forman gradient}. Several algorithms have been developed to compute a Forman gradient, its critical features~\cite{gyulassy2008practical, robins2011theory, shivashankar2011parallel} and simplicial complexes~\cite{weiss2013primal, fellegara2014efficient}. The Forman gradient is the basis for an efficient computation of discrete separatrix lines through mesh traversals. See the supplementary materials for discrete Morse theory details and~\cite{de2015morse} for a comprehensive review.

The Forman method is favored in terrain analysis for its derivative-free nature, offering efficiency and scalability for large datasets~\cite{fellegara2017efficient,fellegara2023terrain}. Yet, its topological accuracy can depend on mesh construction choices, like diagonal selection in grid triangulation~\cite{de2015morse}.
\section{Methodology}\label{sec:method}
\begin{figure*}[h]
    \centering
    \includegraphics[width=0.95\textwidth]{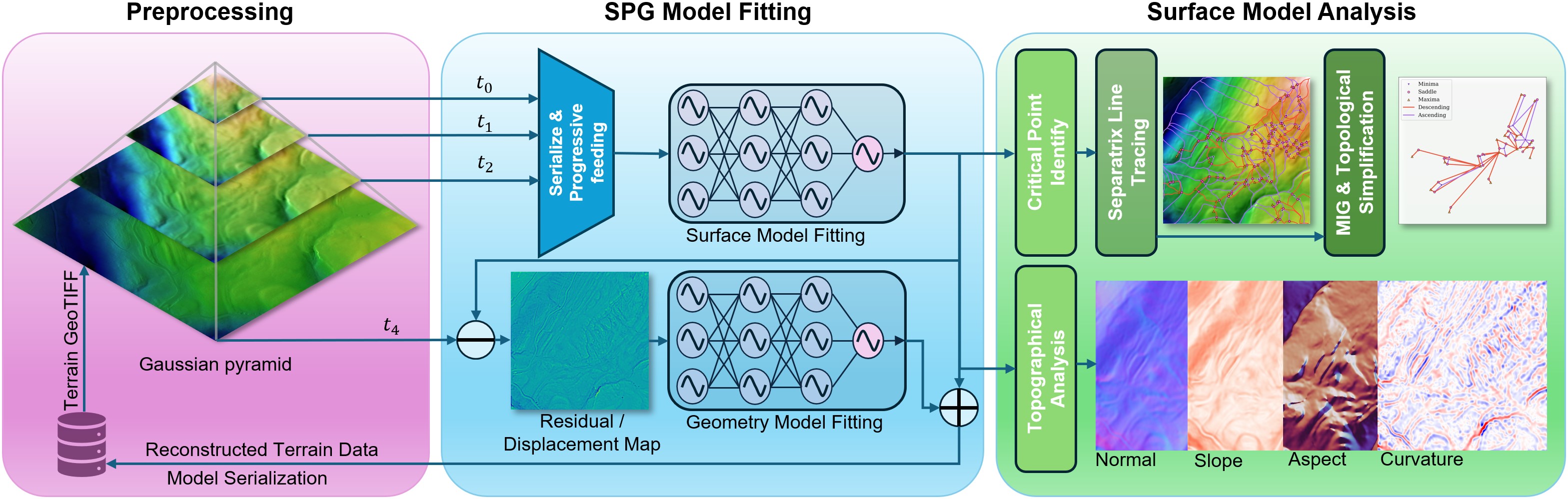}
    \caption{The pipeline of the ImplicitTerrain. Firstly, terrain data is preprocessed as a Gaussian pyramid for progressive fitting. Then, the cascaded Surface-plus-Geometry (SPG) model is trained to fit the smoothed terrain surface and the residual/displacement map in order. Finally, various terrain data analyses are supported by the smooth surface model. Model weights can be serialized for storage and inference to reconstruct the terrain surface with flexible structures (grids or TINs) and resolutions. Better viewed in the digital version.}
    \label{fig:pipeline}
\end{figure*}

ImplicitTerrain represents the terrain surface function $\mathcal{F}_s$ via coordinate-based neural networks that are trained progressively from the ground elevation data $I \in \mathbb{R} ^{H \times W}$ in GeoTIFF~\cite{mahammad2003geotiff} format. The pipeline of ImplicitTerrain, as shown in~\cref{fig:pipeline}, consists of three main stages: (I) \cref{sec:preprocess} Input raster data preprocessing, (II) \cref{sec:fitting} Surface-plus-Geometry (SPG) model fitting, which consists of two cascaded networks: {\em surface model} $\Psi_{s}$ and {\em geometric compensation model} $\Psi_{g}$, and (III) \cref{sec:topology} Topological feature extraction and~\cref{sec:topography} topographical feature computation.

\subsection{Input raster data preprocessing}
\label{sec:preprocess}
\begin{figure}[h]
    \centering
    \includegraphics[width=0.95\columnwidth]{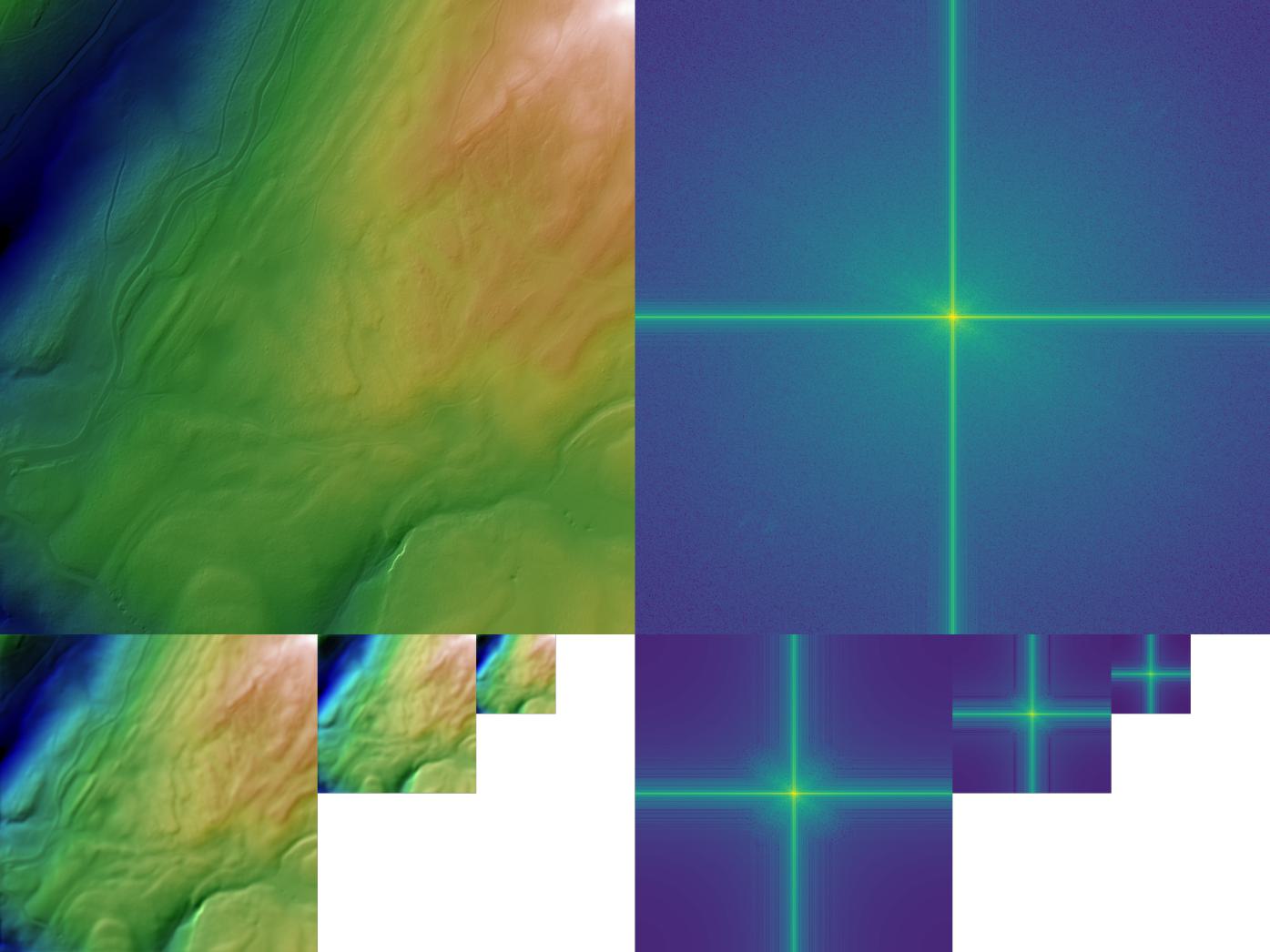}
    \caption{Gaussian pyramid and frequency view of $\text{Swiss}_1$ dataset.}
    \label{fig:gp}
\end{figure}

As a common practice in terrain data analysis, a smoothing process is applied to remove the noise and artifacts during the data acquisition process. This process may diminish the visual quality of the terrain surface since the high-frequency details are absent. To keep the high fidelity of the input signal, we make use of a low-pass Gaussian filter iteratively to construct a multi-resolution Gaussian pyramid of the input raster data as shown in~\cref{fig:pipeline}, where the bottom layer is the original input $I$ and the layer on top is the smoothed version of the previous layer with half the size. This multi-resolution pyramid allows the progressive training of $\Psi_{s}$ from top to bottom (except the bottom layer) as detailed in~\cref{sec:fitting}. As shown in~\cref{fig:gp}, surface details are preserved only in the bottom layer that is used by $\Psi_{g}$.

\subsection{Surface-plus-Geometry (SPG) model fitting}
\label{sec:fitting}
An MLP with sinusoidal non-linear activation function as in SIREN~\cite{sitzmann2020implicit} is used for both $\Psi_{s}$ and $\Psi_{g}$ for simplicity and efficiency. This architecture allows accurate fitting of both $\mathcal{F}_s$ and its derivatives $\{\nabla_{\vec x}\mathcal{F}_s, \nabla_{\vec x}^2\mathcal{F}_s,\ldots \}$. $\Psi_{s}$ and $\Psi_{g}$ can be evaluated at arbitrary query coordinates and derivatives are calculated by the automatic differentiation. The network configuration can be determined heuristically to trade-off between model size and fitting accuracy. However, with the increment of the model size, the training time also increases, especially when $I$ is of high-resolution. Inspired by the {\em spectral bias}, we separate the hard-to-fit high-frequency components from $I$ by the Gaussian pyramid in~\cref{sec:preprocess} and fit $\Psi_{s}$ progressively from top to bottom to reduce the convergence time and smooth the training process. With less computational cost, $\Psi_{s}$ converges faster to the smoothed terrain surface. Then, $\Psi_{g}$ fits the absent details from $\Psi_{s}$ via the residual learning, as shown in~\cref{fig:pipeline}. The residual is the pixel-wise difference between the high-resolution inference of $\Psi_{s}$ and the original input $I$. Fitting the residual shares similar intuition as the {\em displacement map} in computer graphics rendering and in the work by Wang \etal~\cite{yifan2021geometry}. Therefore, $\Psi_{s}$ represents a smooth surface function with a stable gradient field, benefitting the following topological analysis in~\cref{sec:topology}. Compensating $\Psi_{s}$ with surface details in $\Psi_{g}$ via the addition of both models' output, the reconstructed surface function $\hat{\mathcal{F}_s}$ is more accurate and visually plausible than a single model design. As we will show in the ablation study~\cref{sec:ablation}, the SPG model achieves better overall fitting accuracy (PSNR 67.08 dBs \vs 57.66 dBs) and faster convergence speed ($\approx$ 4 times) than a single MLP model with approximately the same number of parameters.

\subsection{Topological feature extraction}
\label{sec:topology}
{\em Surface model} $\Psi_{s}$ can be used as a smooth surface function allowing the direct inference of height values and derivatives at any query locations within the domain. This characteristic greatly simplifies the algorithm design to identify critical points and build the connection between them. Without hindrance, Morse theory can be directly applied to the smooth function $\Psi_{s}$, avoiding the difficulties of approximating the gradient field on discrete meshes.

\textbf{Identify critical points.} Critical points are the points where the gradient vanishes (\ie, $||\nabla\mathcal{F}_s||=0$). For terrain surface, critical points can be classified into three types: {\em Maximum}, {\em Minimum}, and {\em Saddle}. According to Sylvester's criterion~\cite{gilbert1991positive}, this classification can be achieved by the second-order derivatives, \ie, Hessian matrix $\mathcal{H}|_{\vec{p}} \in \mathbb{R}^{2\times 2}$ at the critical point coordinate $\vec{p}$. To locate the critical points, we trace the zero-cross points of the gradient field by a simple gradient descent algorithm to minimize the gradient norm $||\nabla \Psi_s||$. Candidate critical point locations are determined by considering the value differences of the neighboring vertices to the center vertex on smoothed meshes. Clone and random jitter of the coordinates are applied to these candidate coordinates. For each step, the gradient norm $||\nabla \Psi_s||$ minimization direction $\vec{d}$ takes the form:
\begin{equation}
    \begin{split}
        \vec{d} =& -\left\langle \partial ||\nabla \Psi_s||/\partial x, \partial ||\nabla \Psi_s||/\partial y\right\rangle \\ =& -\left\langle f_x f_{xx} + f_y f_{yx},f_{x} f_{xy} + f_y f_{yy} \right\rangle
    \end{split}
\end{equation}
where $f_x, f_y$ are the first-order derivatives of $\Psi_s$ and $f_{xx}, f_{xy}, f_{yx}, f_{yy}$ are the second-order derivatives.

\textbf{Separatrix lines and MIG.} As detailed in~\cref{sec:tda}, separatrix lines are a subset of integral lines that connect the critical points, depicting the topological skeleton of the terrain surface. For each {\em Saddle}, the separatrix lines connect it to two neighboring {\em Minimum} and two {\em Maximum} points. The eigenvector of the Hessian matrix $\mathcal{H}$ at the {\em Saddle} point is used to determine the starting directions of the four separatrix lines. Heading to the {\em Maxima}/{\em Minima}, the gradient field is traced to maximize/minimize the height values. Determining the relationship between the critical points helps to build the MIG of full resolution, from which, through persistence-based simplification, the prominent topological structure of the terrain surface is extracted as in~\cref{fig:pipeline}. The design and implementation details of the tracing algorithms are provided in the supplementary material.

\section{Experiments}\label{sec:exp}
We evaluate the performance of ImplicitTerrain on terrain dataset from two aspects: model accuracy and topological analysis results. We employ an open-source library\footnote{Publicly accessible at \url{https://github.com/UMDGeoVis/FormanGradient2D}} 
to compute the separatrix lines and simplified MIGs by computing Forman gradients. In the rest of this paper, we refer to this library as the {\em Forman method}. Due to the lack of ground truth topology of the real-world terrain, in~\cref{sec:synthetic}, a synthetic terrain model is generated to demonstrate the effective alignment between results from ImplicitTerrain and {\em Forman method}.  In~\cref{sec:real}, high-resolution terrain data is fed to the pipeline. In~\cref{sec:ablation}, we design an ablation experiment to show the effectiveness and efficiency of our SPG model in terms of training time and fitting results. In~\cref{sec:noise}, we conduct experiments for the noise robustness of the surface topology obtained from our surface model.

\subsection{Experiment setup and evaluation metrics}\label{sec:exp_setup}
Both $\Psi_{s}$ and $\Psi_{g}$ use an MLP with 3 hidden layers and 256 hidden units per layer to fit an input 1000$\times$1000 raster data $I$ for the height field of $1\text{km}^2$ terrain surface. Height values and the coordinates are normalized into the $[-1, 1]$ range. The sinusoidal activation function and the same weight initialization are applied as in SIREN~\cite{sitzmann2020implicit}. As in~\cref{fig:pipeline}, $\Psi_{s}$ is progressively fitted on the top 3 layers of a 4-layer Gaussian pyramid constructed with a Gaussian kernel size $\sigma=4.0$. The final optimization target $I_s$ of $\Psi_{s}$ is of size 500$\times$500. Then, $\Psi_{g}$ fits the residual between $I$ and the surface model's inference output $I_{\Psi}$. Both models are trained for 3000 steps with the Adam optimizer with a learning rate of 1e-4. All the experiments are conducted on a computation node with a 24-core CPU (AMD EPYC 7352) and an NVIDIA A5000 GPU. From the aspects of model accuracy and topological analysis results, various evaluation metrics are used to measure the performance of the ImplicitTerrain.

\textbf{Model accuracy.} To measure the reconstruction accuracy of the ImplicitTerrain, peak signal-to-noise ratio (PSNR) and structural similarity index measure (SSIM) are calculated. For PSNR calculation, since there is no boundary for the height field, the maximum possible value of the signal is set to $1.0$. SSIM metric evaluates the visual similarity between the model output and the original input data. For evaluating the surface model $\Psi_{s}$, the frequency band reconstruction capability is calculated by the difference between the log-scaled magnitude of the Fourier transform of $I_s$ and $\Psi_{s}$'s inference result in $I_{\Psi}$, \ie $\log(|I_{s}|) - \log(|I_{\Psi}|)$. Since the gradient field is important for topological and topographical analysis, we measure the gradient field reconstruction accuracy in terms of the pixel-wise difference of gradient norm and direction between our model's estimations $\nabla \Psi_{s}$ and the results derived from $I_s$ by pixel shifting.

\textbf{Topological analysis.} To evaluate the topological analysis result of $\Psi_{s}$, we measure the degree of alignment between the topology of our surface model and the topology from {\em Forman method}. The MIGs from both methods are simplified by a persistence value of 1 meter which is heuristically set to keep prominent critical points from {\em Forman method}. The MIG alignment is measured from two aspects: (1) critical points matching as a binary classification method, and (2) MIG-based persistence diagram. Critical points are matched based on spatial closeness and the result is reported in {\em Precision, Recall}, and {\em $F_{0.5}$ score}, which regards the precision as twice as important as the recall rate. To highlight the topological features in the MIG, we make use of its persistence diagram, which is a visualization of connected critical point pairs~\cite{fugacci2016persistent}. It depicts the evolution of the terrain surface topology and highlights the critical points representing prominent topological features, \eg, mountain peaks. We use the Wasserstein distance ($WS$) to measure the similarity between two persistence diagrams, which is defined as the minimal distance achieved by a perfect matching of two diagrams. A smaller $WS$ value indicates a higher similarity between two persistence diagrams.

For the {\em Forman method} applied to the regular gridded mesh, it is known that the {\em mesh configuration} (connectivity between neighboring vertices in a diagonal direction) may influence the MIG result~\cite{de2015morse}. Satisfying the Delaunay triangluation~\cite{delaunay1934sphere} of the gridded data, we consider two different mesh configurations for the {\em Forman method}: each vertice is connected to its (N, S, E, W, NW, SE) or (N, S, E, W, NE, SW) neighbors, named as {\em Mesh1} and {\em Mesh2} respectively. Considering the missing ground-truth persistence diagram and the influence of mesh configuration on the MIG, we further propose to evaluate the MIGs by the ratio of Wasserstein distance as
\begin{equation}
    WS_{ratio} = \frac{\min(WS(\Psi_{s}, \text{Mesh1}), WS(\Psi_{s}, \text{Mesh2}))}{WS(\text{Mesh1}, \text{Mesh2})}
\end{equation}
If this ratio value is between 0 and 1, it means that the MIG from $\Psi_{s}$ is closer to the {\em Forman method} result than the distance between the MIGs of {\em Mesh1} and {\em Mesh2}. Thus, we consider the topology obtained from our method to be accurate. Limited by the paper length, details about the Wasserstein distance calculation and the influence of mesh configuration are provided in the supplementary materials.
\subsection{Synthetic terrain data}\label{sec:synthetic}
\begin{figure}[t]
    \centering
    \begin{subfigure}{0.45\linewidth}
        \includegraphics[width=\linewidth]{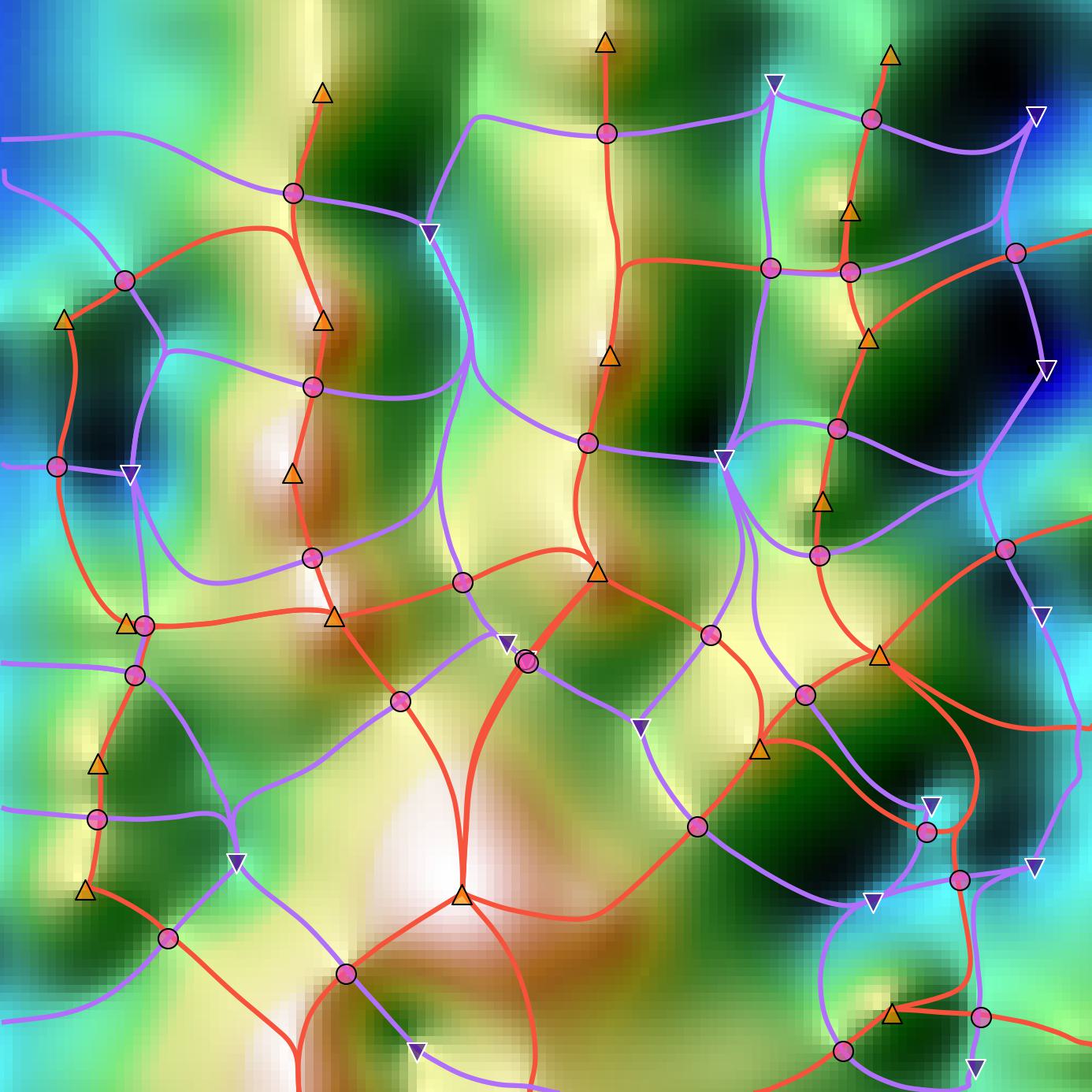}
        \caption{Separatrix lines - ImplicitTerrain.}
        \label{fig:art_sep_inr}
    \end{subfigure}
    \hfill
    \begin{subfigure}{0.45\linewidth}
        \includegraphics[width=\linewidth]{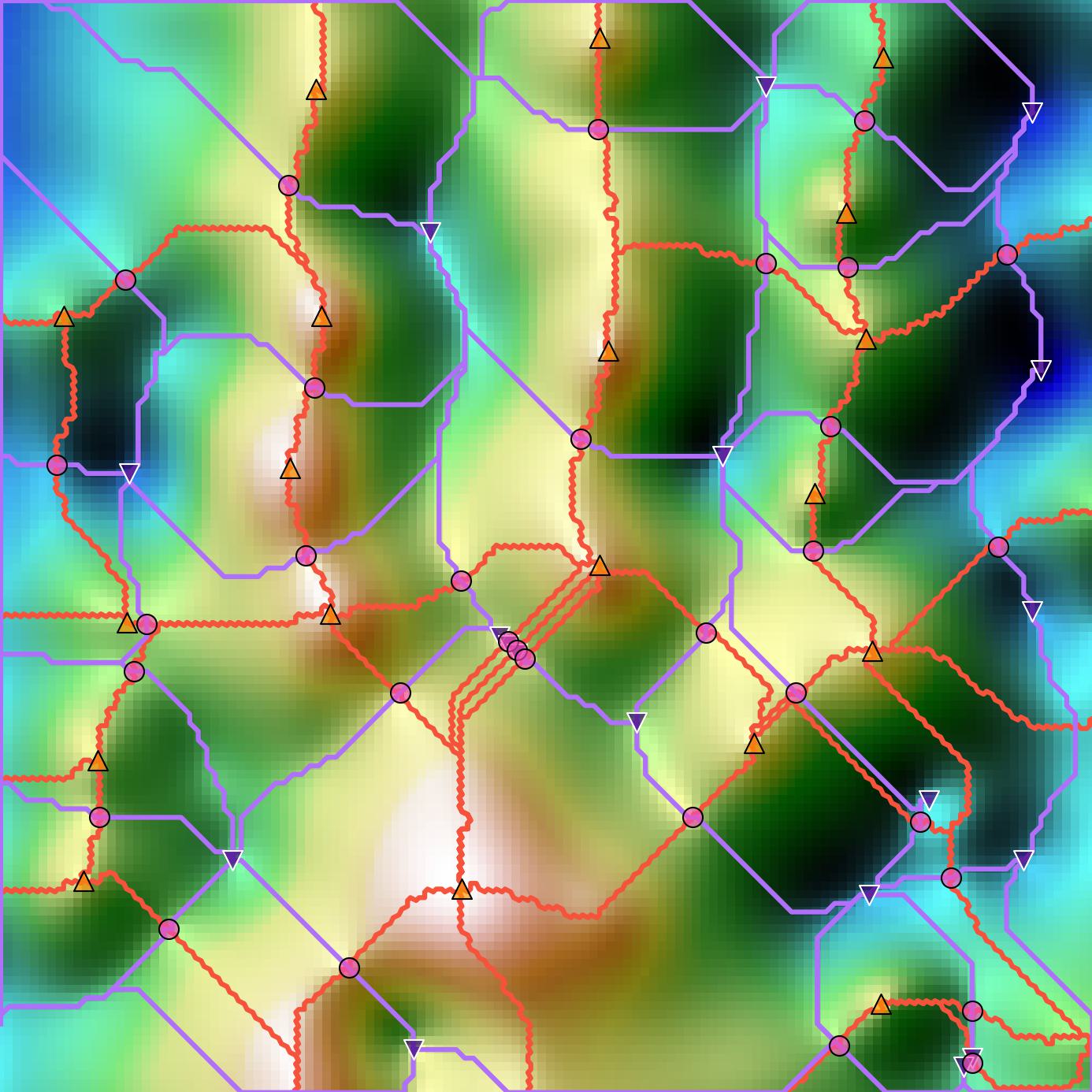}
        \caption{Separatrix lines - {\em Forman method}.}
        \label{fig:art_sep_forman}
    \end{subfigure}
    \begin{subfigure}{0.45\linewidth}
        \frame{\includegraphics[width=\linewidth]{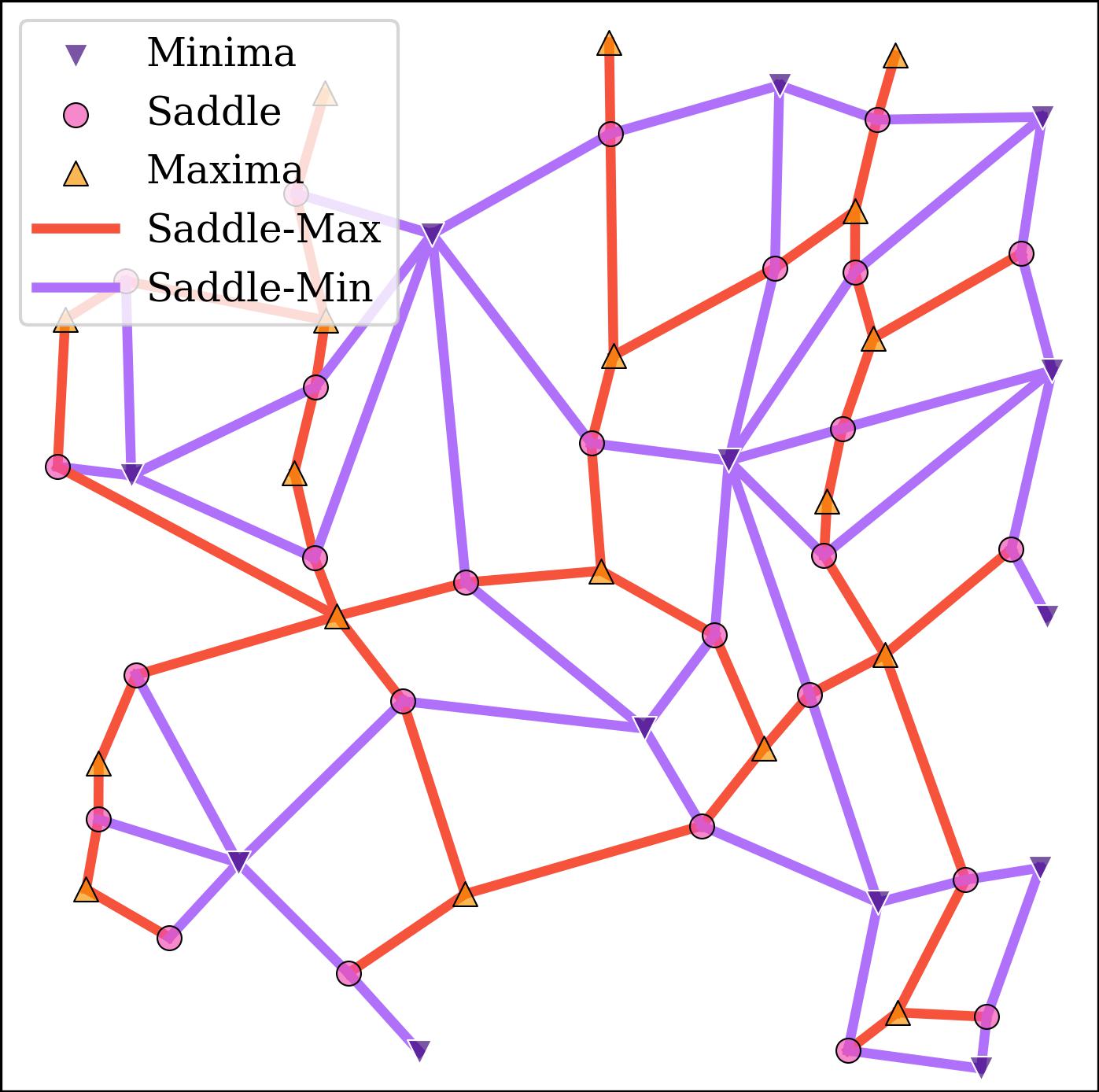}}
        \caption{MIG - ImplicitTerrain.}
        \label{fig:art_mig_inr}
    \end{subfigure}
    \hfill
    \begin{subfigure}{0.45\linewidth}
        \frame{\includegraphics[width=\linewidth]{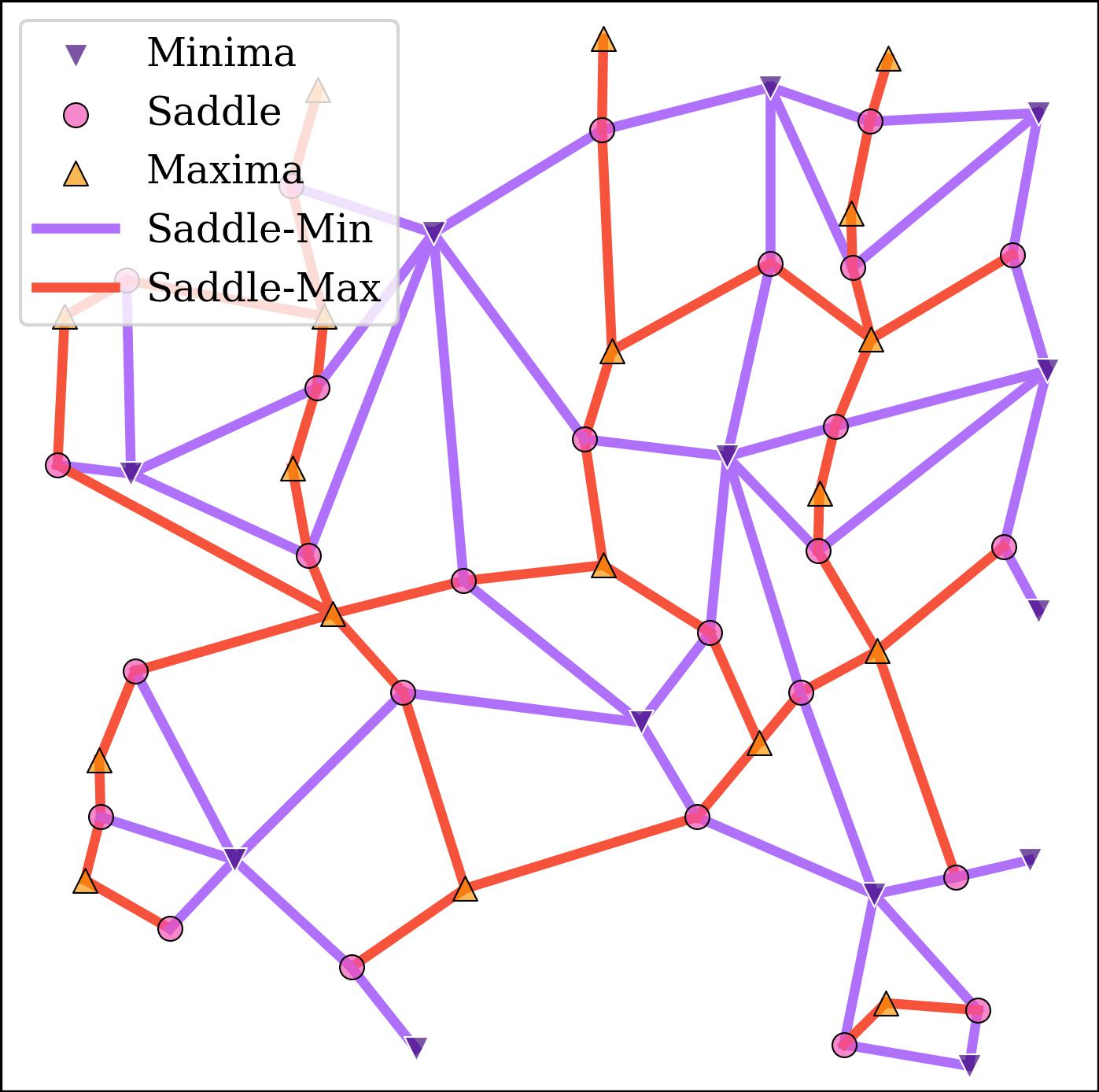}}
        \caption{MIG - {\em Forman method}.}
        \label{fig:art_mig_forman}
    \end{subfigure}
    \caption{Comparison of topological analysis results of the synthetic terrain. Node colors and shapes represent the critical point types and the edge colors represent the separatrix lines as in the legend of (c) and (d). Better viewed in the digital version.}
    \label{fig:art}
\end{figure}

Before evaluating the ImplicitTerrain on real-world terrain data, synthetic terrain data with simple topology is generated to show the effective alignment of our surface model analysis results to the {\em Forman method} results. The synthetic terrain is of side length 256 with height values between $[0, 200]$ and is generated as a combination of 2D random Gaussian signals with various covariances as shown in~\cref{fig:f1}. Particularly, knowing the surface gradient of the synthetic terrain allows a better data triangulation than the {\em Mesh1} and {\em Mesh2}. The triangulation of each grid cell is determined by the surface gradient, leading to visually better separatrix lines by {\em Forman method}. Due to the simplicity of this dataset, our models are also shrunk to have 3 hidden layers and 128 hidden units. As shown in~\cref{fig:art}, the critical points, separatrix lines, and MIGs from both methods form a good visual alignment. Quantitative evaluations of the topological analysis results are shown in~\cref{tab:topo}. $F_{0.5}$-score value is $1.0$ and the small $WS_{ratio}$ proof the high similarity between our results and {\em Forman method}. Besides the correctness of the topological analysis results, it is obvious that the smooth (zigzag free) separatrix lines obtained from the ImplicitTerrain are visually plausible and faithfully align with the terrain surface. These smooth and accurate separatrix lines might lead to better segmentation results based on the Morse-Smale Complexes.

\begin{table}
    \centering
    \begin{tabular}{@{}lccccc@{}}
        \toprule
        {\bf Name}            & {\bf precision} & {\bf recall} & {\bf $F_{0.5}$ score} & $WS_{ratio}$ \\
        \midrule
        $\text{Synth}_{ours}$ & 1.00            & 1.00         & 1.00                  & 0.68         \\
        \midrule
        $\text{Swiss}_1$      & 0.90            & 0.96         & 0.91                  & 0.17         \\
        $\text{Swiss}_2$      & 0.91            & 0.831        & 0.89                  & 0.31         \\
        $\text{Swiss}_3$      & 0.89            & 0.78         & 0.87                  & 0.69         \\
        $\text{Swiss}_4$      & 0.91            & 0.83         & 0.89                  & 0.35         \\
        \bottomrule
    \end{tabular}
    \caption{Topological analysis results of the synthetic and real-world terrain. $WS_{ratio}$ between $[0, 1]$ indicates the MIGs from both methods are well aligned.}
    \label{tab:topo}
\end{table}

\subsection{Real-world terrain data}\label{sec:real}
\begin{figure}[t]
    \centering
    \begin{subfigure}{0.45\linewidth}
        \includegraphics[width=\linewidth]{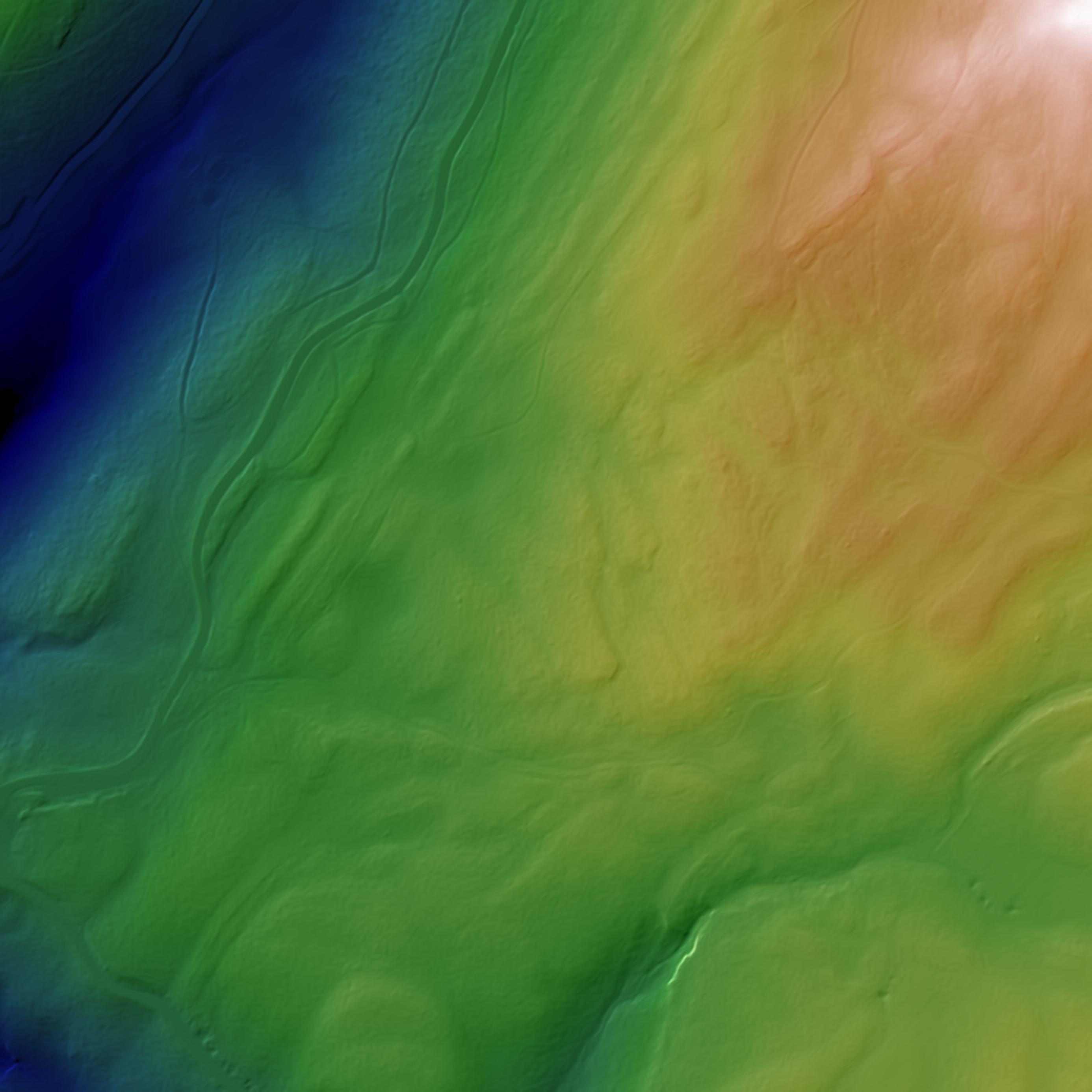}
        \caption{Hillshade of SPG output.}
        \label{fig:s1_input}
    \end{subfigure}
    \hfill
    \begin{subfigure}{0.45\linewidth}
        \includegraphics[width=\linewidth]{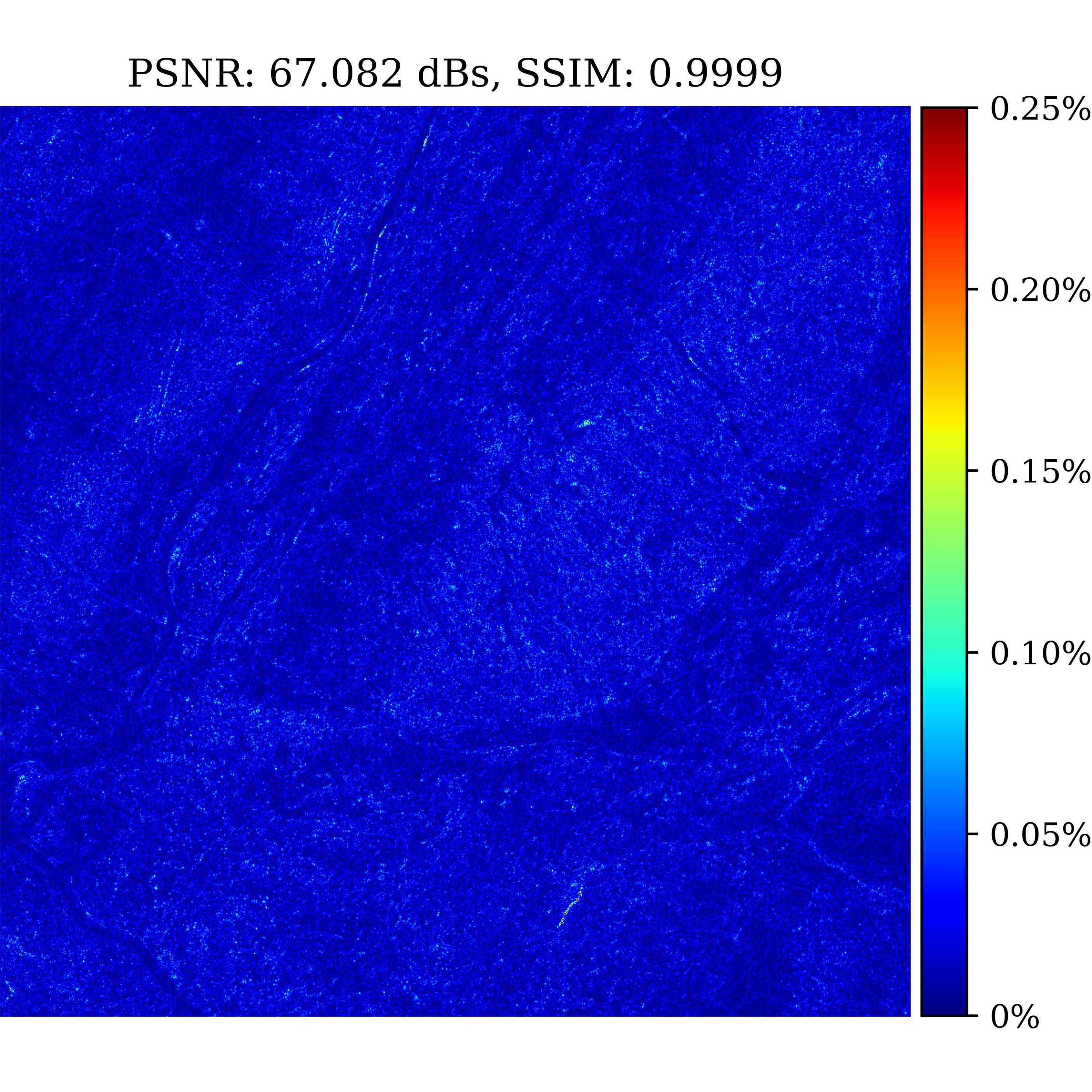}
        \caption{Absolute error of the fitting.}
        \label{fig:s1_inr}
    \end{subfigure}
    \begin{subfigure}{0.45\linewidth}
        \includegraphics[width=\linewidth]{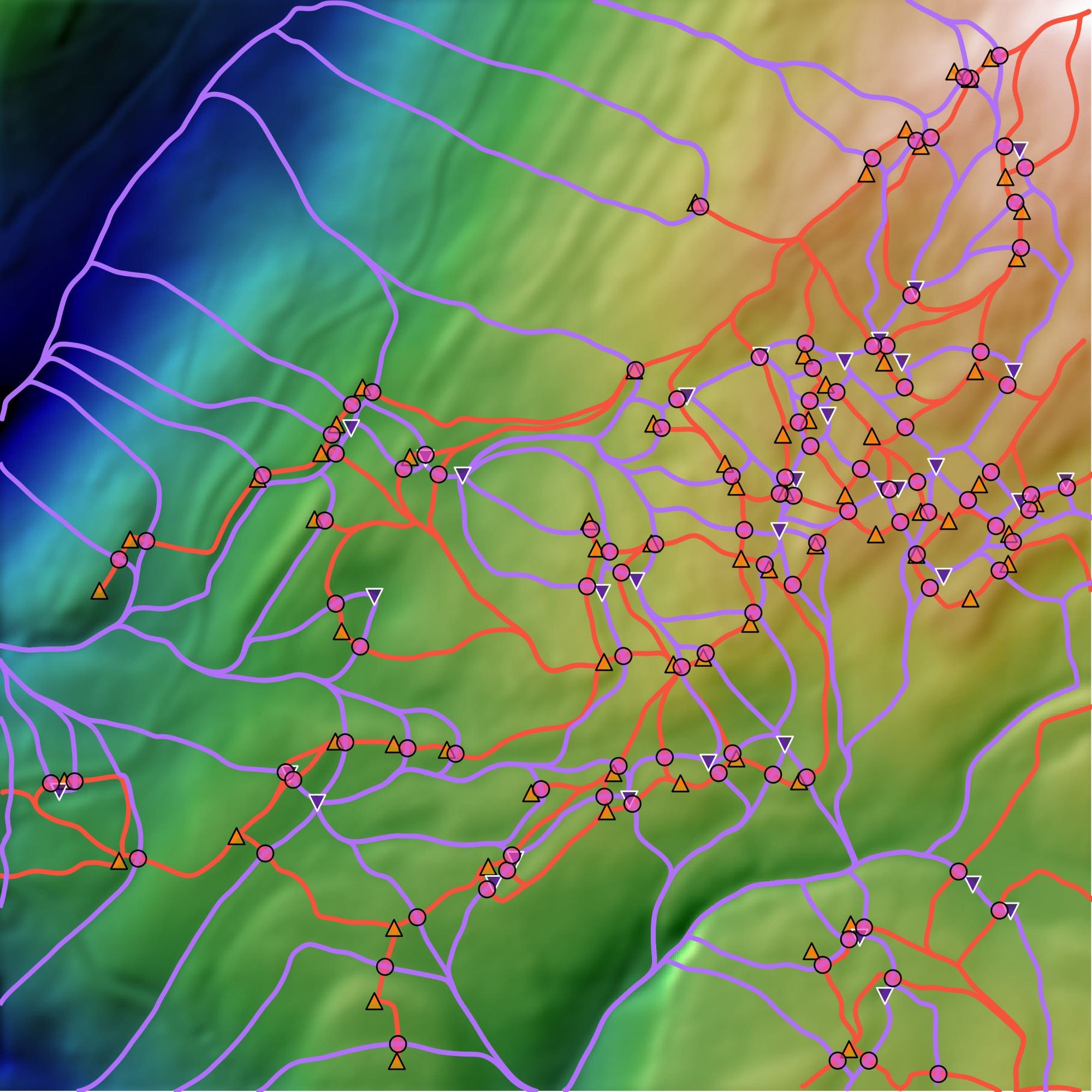}
        \caption{Separatrix lines.}
        \label{fig:s1_separatrix}
    \end{subfigure}
    \hfill
    \begin{subfigure}{0.45\linewidth}
        \frame{\includegraphics[width=\linewidth]{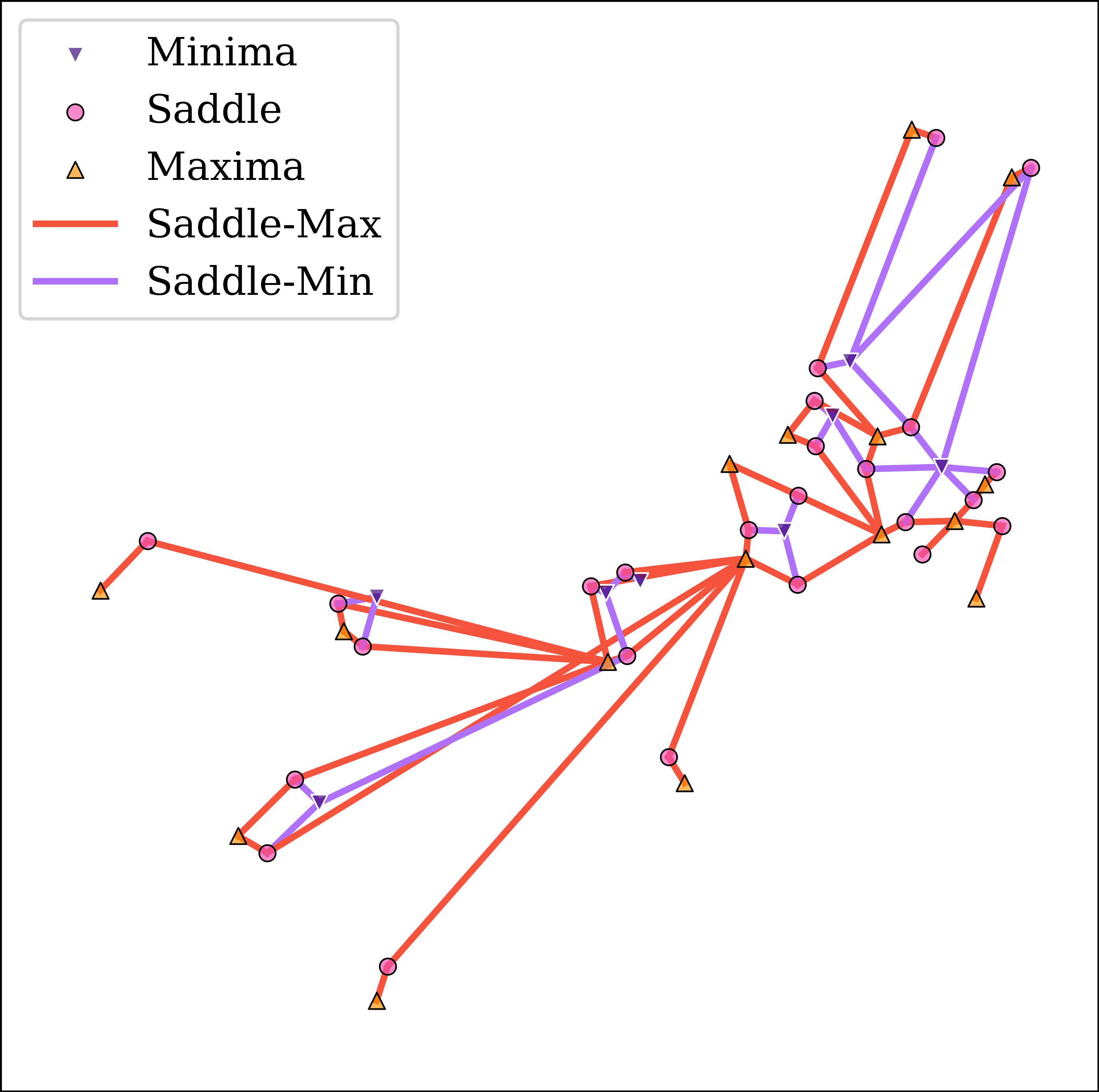}}
        \caption{MIG.}
        \label{fig:s1_mig}
    \end{subfigure}
    \caption{ImplicitTerrain fitting results and topological features extracted from $\text{Swiss}_1$. (b) shows the fitting error with red color mapping to {\bf 0.25\%} error. Better viewed in the digital version.}
    \label{fig:s1}
\end{figure}

Swisstopo dataset {\em swissALTI3D}~\cite{swissALTI3D} provides high-resolution digital elevation information of the Switzerland terrain surface without vegetation. Four different tiles $\text{Swiss}_k$ ($k \in \{1,2,3,4\}$) of size 1km$\times$1km are selected with various terrain characteristics as input data and each tile is of resolution 1 meter per pixel. Limited by the paper length, we only put the results for $\text{Swiss}_1$ tile and the rest results with extra analysis can be found in the supplementary materials. As shown in~\cref{tab:swiss_fit}, ImplicitTerrain model size is 80\% less than the input raster file and the surface fitting criteria all show our method provides a very good reconstruction of the terrain surface function. The gradient field, especially the gradient direction, is accurately reconstructed, benefitting the following topological analysis and topographical calculation. Besides, in~\cref{fig:s1}, the detailed textures over the terrain surface are preserved. With the accurate fitting of the terrain gradient field, in~\cref{tab:topo}, the results of critical point matching and Wasserstein distance ratio show that our surface model is able to correctly capture the topological features from the input terrain. The value of the Wasserstein distance ratio is small enough to confirm that our persistence diagram is closely aligned with the {\em Forman method} result. As shown in~\cref{fig:s1}, separatrix lines and critical points together delineate the terrain topology. Ridges/valleys between the mountain saddles and the peaks/basins are accurately depicted in the plot.
\begin{table*}[t]
    \centering
    \begin{tabular}{|l|p{1.5cm}|p{1cm}|p{1cm}|p{1cm}|p{1.5cm}|p{1.5cm}|p{1.8cm}|p{1cm}|p{1cm}|}
        \hline
        {\bf Name}       & {\bf Sizes (MBs)} & {\bf Size ratio} & {\bf $\Psi_{s}$ PSNR} & {\bf $\Psi_{s}$ SSIM} & {\bf Freq diff} $\times 10$ & {\bf Grad norm diff} $\times 10$ & {\bf Grad direction diff (rad)} $\times 10$ & {\bf SPG PSNR} & {\bf SPG SSIM} \\
        \hline
        $\text{Swiss}_1$ & $1.51/7.6$        & 0.20             & 64.85                 & 0.9999                & 1.49$\pm$2.31               & 0.54$\pm$0.52                    & 0.62$\pm$1.10                               & 67.08          & 0.9999         \\
        $\text{Swiss}_2$ & $1.51/7.6$        & 0.20             & 60.53                 & 0.9998                & 0.95$\pm$2.08               & 0.77$\pm$1.00                    & 0.61$\pm$0.77                               & 52.34          & 0.9992         \\
        $\text{Swiss}_3$ & $1.51/7.6$        & 0.20             & 59.75                 & 0.9998                & 0.13$\pm$0.29               & 0.86$\pm$1.05                    & 0.72$\pm$1.02                               & 58.93          & 0.9997         \\
        $\text{Swiss}_3$ & $1.51/7.6$        & 0.20             & 62.54                 & 0.9999                & 0.17$\pm$0.32               & 0.56$\pm$0.61                    & 0.46$\pm$0.57                               & 66.59          & 0.9999         \\
        \hline
    \end{tabular}
    \caption{Numerical evaluation of the fitting results of the real-world terrain. {\bf Sizes} are the total model sizes and the input raster size, and {\bf Size ratio} is their ratio. {\bf $\Psi_{s}$ PSNR} and {\bf $\Psi_{s}$ SSIM} are the fitting accuracy of the surface model to the smoothed data. {\bf SPG PSNR} and {\bf SPG SSIM} are the fitting accuracy of the SPG model to the original input. For the surface model, {\bf Freq diff} is the mean and standard deviation of the frequency domain difference. {\bf Grad norm/direction diff}s are the mean and standard deviation of the difference of gradient norm and direction between $\nabla \Psi_{s}$ and the estimated image gradient from $I_s$. $\times 10$ denotes the scaling factor for better numerical representation.}
    \label{tab:swiss_fit}
\end{table*}

\subsection{Ablation study: SPG model \vs single model}\label{sec:ablation}
\begin{figure}[t]
    \centering
    \includegraphics[width=0.95\columnwidth]{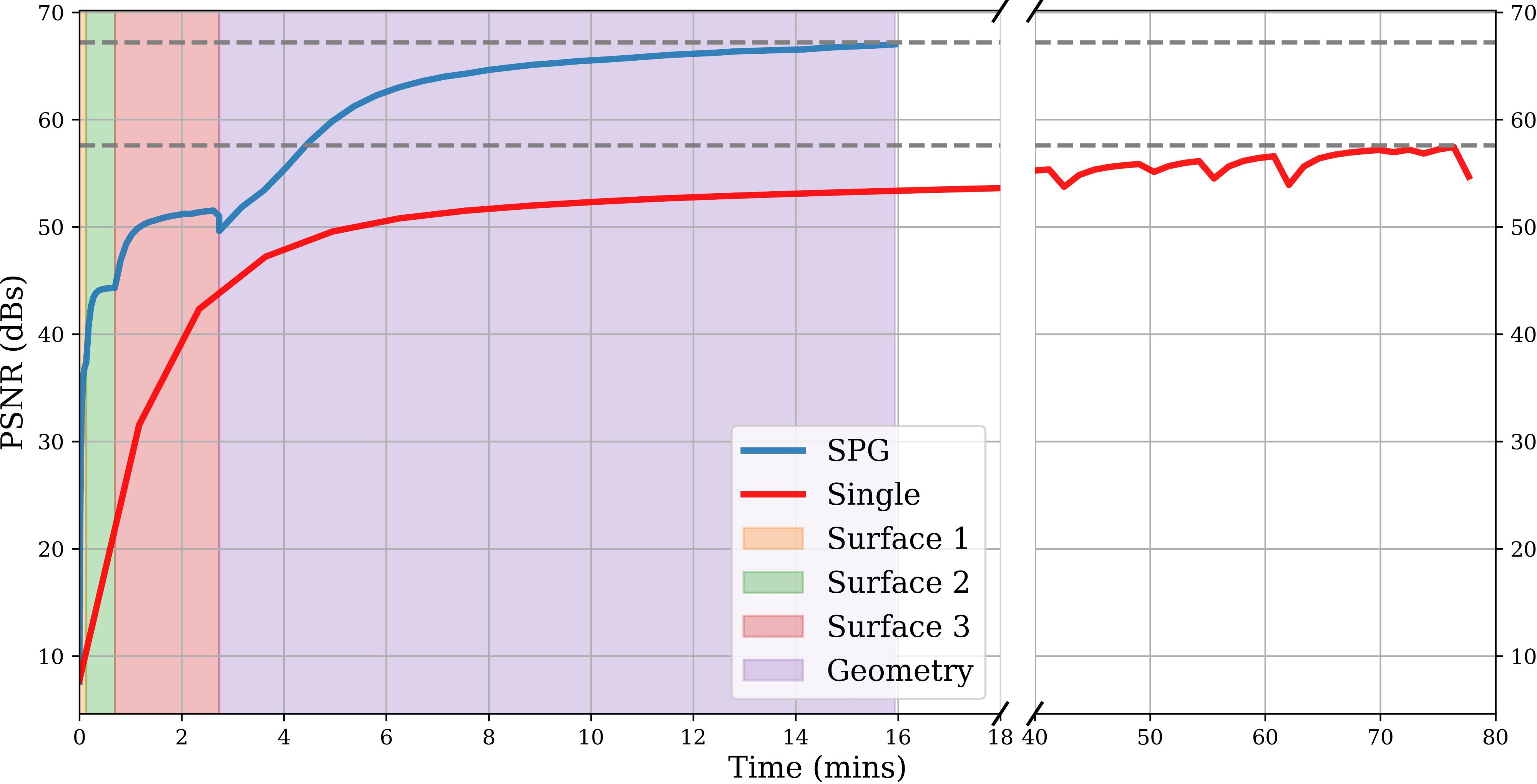}
    \caption{Ablation comparison of the fitting accuracy between SPG and single model \wrt training time. Colored regions denote different training stages of the SPG model.}
    \label{fig:ablation_time}
\end{figure}

\begin{figure}[t]
    \centering
    \begin{subfigure}{0.49\linewidth}
        \includegraphics[width=\linewidth]{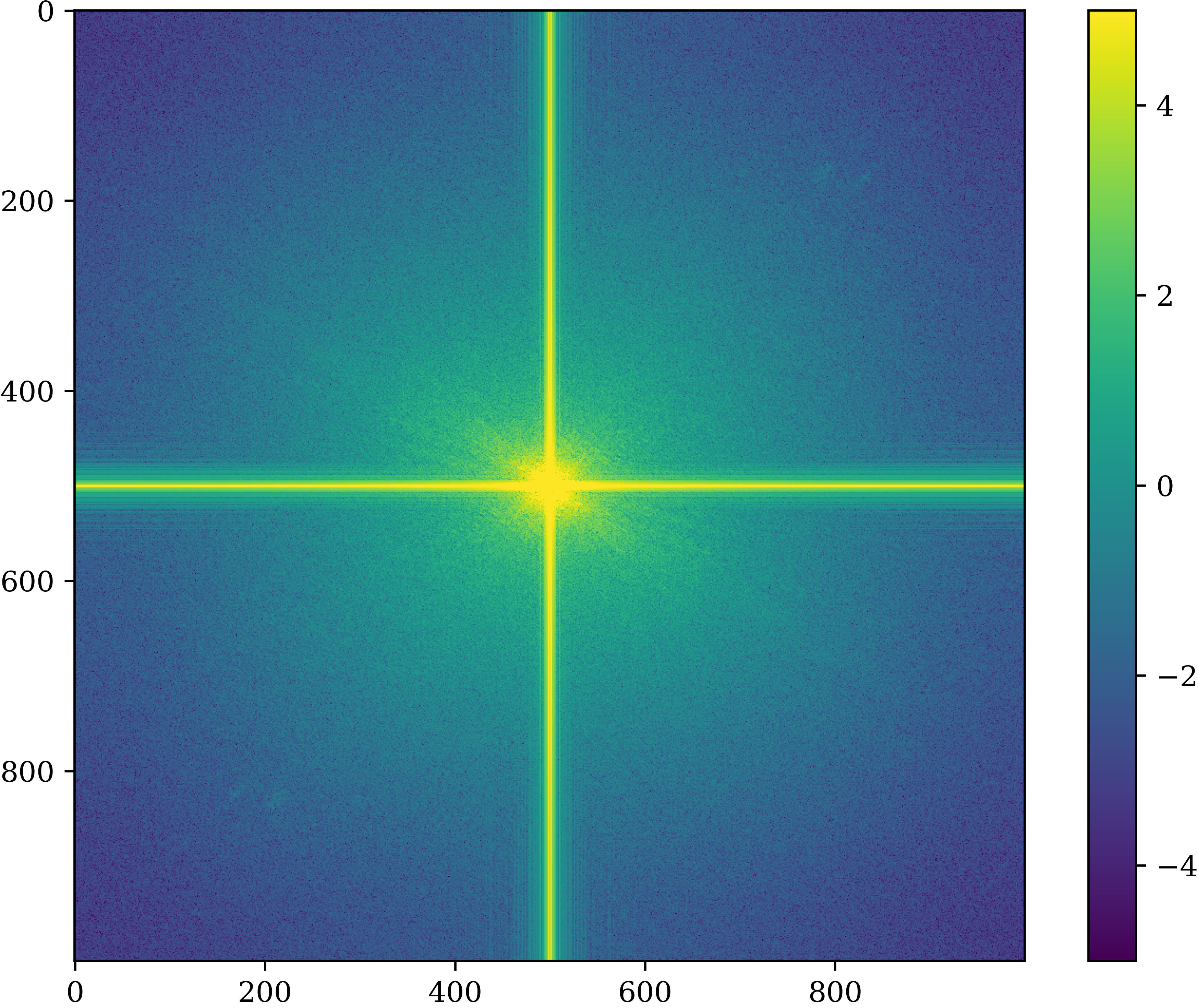}
        \caption{Fourier transform of the $\text{Swiss}_1$.}
        \label{fig:ab_ref_fft}
    \end{subfigure}
    \hfill
    \begin{subfigure}{0.49\linewidth}
        \includegraphics[width=\linewidth]{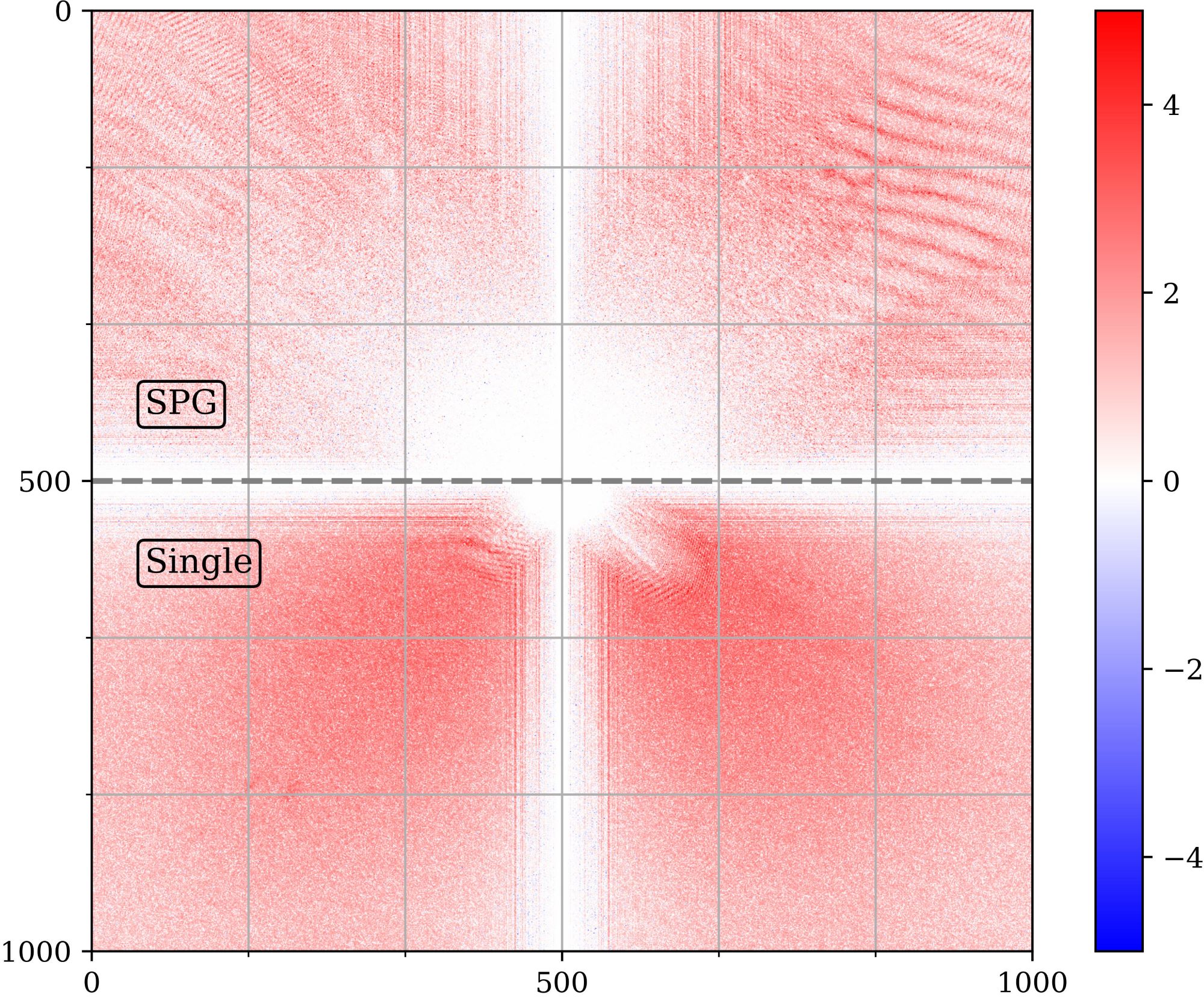}
        \caption{Frequency domain loss.}
        \label{fig:ab_fft_diff}
    \end{subfigure}
    \caption{Freqency domain comparison of the fitting results. (a) shows the ground truth. (b) Since the Fourier transform result is central symmetric, SPG and single model frequency domain loss are plotted together as labeled in the figure.}
    \label{fig:ablation_fft}
\end{figure}

To counter the harm of {\em spectral bias}, we propose this simple-but-effective SPG pipeline which takes advantage of the downstream task requirement - it is a common practice to smooth the terrain surface before topological and topographical analysis, especially for high-resolution inputs. Our SPG model preserves a detailed visual quality of the terrain surface texture while effectively capturing topological features as in~\cref{fig:s1}. To further show this design's benefit to fitting accuracy and training time, we compare it to a single model with about the same number of trainable parameters - 3 hidden layers with 512 hidden units. In~\cref{fig:ablation_time}, we plot the smoothed (via Exponential Moving Average) PSNR values averaged for 5 rounds, with reference to the training time. Same as in the pipeline, both models are fitted for 6000 steps in total. Our progressively fitting model converges much faster and more accurately than the single model which takes {\bf 4} times more time to achieve a worse fitting accuracy (PSNR 67 dBs \vs 58 dBs). Additionally, by analyzing the frequency domain difference, \Cref{fig:ablation_fft} shows that the design of SPG helps increase the fitting capacity of the network, explaining the better PSNR value it achieves.

\subsection{Noise robustness}\label{sec:noise}
\begin{figure}[t]
    \centering
    \includegraphics[width=0.85\columnwidth]{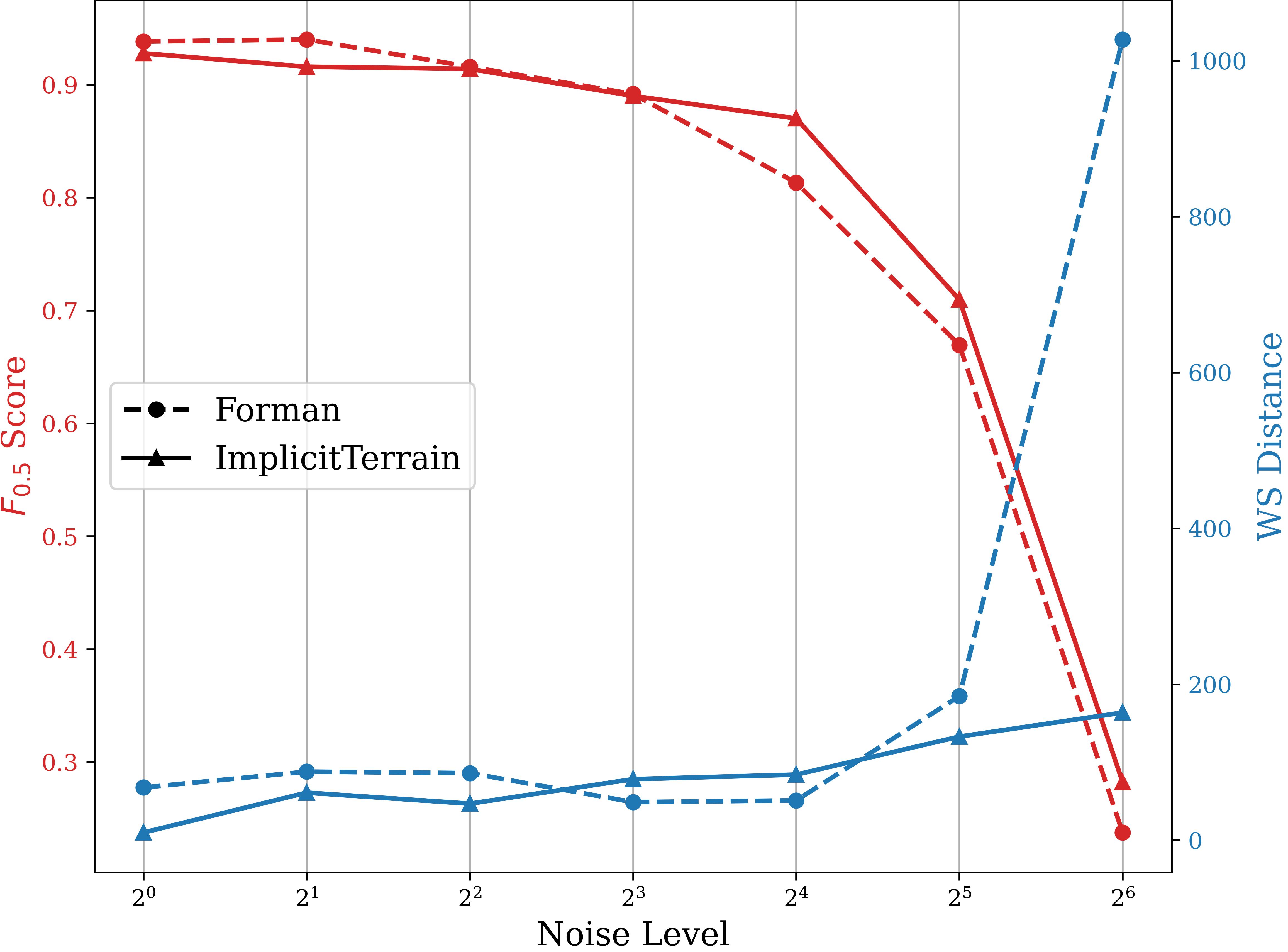}
    \caption{Comparison of noise robustness. {\em Forman method} and ImplicitTerrain comparison via $F_{0.5}$ score and Wasserstein distance of the $\text{Swiss}_1$ \wrt noise level.}
    \label{fig:noiserobust}
\end{figure}

To evaluate the model's robustness to the presence of noise, pixel-wise Gaussian random noise is added to the normalized input with the variance of noise growing from 1e-3 to 6.4e-2 exponentially. Averaging the results from 5 rounds, \Cref{fig:noiserobust} shows the changes of $\text{Swiss}_1$'s $F_{0.5}$-scores and Wasserstein distance to the reference (original input) with the increment of noise level. ImplicitTerrain achieves on par result with the {\em Forman method} at the beginning and shows superior robustness when the noise level further increases.

\subsection{Terrain topographical analysis}
\label{sec:topography}
Our ImplicitTerrain, as a smooth model of the terrain surface, can easily support the topographical analysis that requires the computation of surface derivatives~\cite{mitavsova1993interpolation}. Since the surface function $F(x,y)=z$ is a scalar function defined over the 2D domain, the normal direction can be derived as $\left<-f_x, -f_y, 1\right>$, where $f_x, f_y$ are the first-order derivatives of the function. According to~\cite{o2006elementary}, the mean curvature of the manifold can be derived from the second-order derivative of the function as
\begin{equation}
    H = \frac{(1 + f_y^2)f_{xx} - 2f_xf_yf_{xy} + (1 + f_x^2)f_{yy}}{2(1+f_x^2+f_y^2)^{3/2}}
\end{equation}
where $f_{xx}, f_{xy}, f_{yy}$ are the second-order derivatives of the function. In the last stage of~\cref{fig:pipeline}, we show the basic topographical terrain attributes of $\text{Swiss}_1$ derived from the gradient of $\Psi_{s}$, \ie normal map, slope, aspect, and mean curvature. Limited by the paper length, detailed definition, and high-resolution results in the supplementary materials.

\section{Conclusion}
\label{sec:conclusion}
In this paper, we propose an ImplicitTerrain pipeline to accurately model high-resolution terrain data and support various downstream analyses over a smooth surface representation. Digital surface model and geometric details are sequentially learned by our Surface-plus-Geometry (SPG) model, in which the surface model fitting is accelerated by a multi-stage learning strategy. Furthermore, the smooth and accurate terrain model enables Morse theory-based topological analysis results to be reliable and interpretable. In the future, we plan to process high-resolution point clouds directly and extend the pipeline to support large-scale terrain regions with better model design. The training time and inferencing computation burden can be further optimized by the meta-learning and model quantization correspondingly. We believe that ImplicitTerrain will benefit many research fields that require accurate surface modeling and topological data analysis.

\section*{Acknowledgment}
This work has been supported by the US National Science Foundation under grant number IIS-1910766.

{
    \small
    \bibliographystyle{ieeenat_fullname}
    \bibliography{main}
}

\end{document}